\documentclass[letterpaper]{article}

\usepackage{enumitem}
\usepackage{ragged2e}
\usepackage{colortbl} 
\usepackage{arydshln}
\usepackage{amsmath}
\usepackage[d]{esvect}
\usepackage{booktabs}
\usepackage{hyperref}

\usepackage{tikz}
\usepackage{multirow}

\usepackage{aaai24}  
\usepackage{times}  
\usepackage{helvet}  
\usepackage{courier}  
\usepackage{graphicx} 
\usepackage{natbib}  
\usepackage{caption} 
\usepackage[most]{tcolorbox}
\usepackage{algorithm}
\usepackage{algorithmic}

\usepackage{newfloat}
\usepackage{listings}
\DeclareCaptionStyle{ruled}{labelfont=normalfont,labelsep=colon,strut=off} 
\floatstyle{ruled}
\newfloat{listing}{tb}{lst}{}
\floatname{listing}{Listing}
\title{Zero-Shot Stance Detection using Contextual Data Generation with LLMs}
\author{
   Ghazaleh Mahmoudi, 
   Babak Behkamkia, 
   Sauleh Eetemadi 
}
\affiliations{
     School of Computer Engineering,
Iran University of Science and Technology, Iran\\
       \texttt{\{gh\_mahmoudi, babak\_behkamkia\}@comp.iust.ac.ir}, 
       \texttt{sauleh@iust.ac.ir}

}

\usepackage{bibentry}

\begin{document}

\maketitle

\begin{abstract}
Stance detection, the classification of attitudes expressed in a text towards a specific topic, is vital for applications like fake news detection and opinion mining. However, the scarcity of labeled data remains a challenge for this task. To address this problem, we propose \textbf{Dy}namic \textbf{M}\textbf{o}del \textbf{Adap}tation with Contextual Data Generation (DyMoAdapt) that combines Few-Shot Learning and Large Language Models. In this approach, we aim to fine-tune an existing model at test time. We achieve this by generating new topic-specific data using GPT-3. 
This method could enhance performance by allowing the adaptation of the model to new topics. However, the results did not increase as we expected. Furthermore, we introduce the 
\textbf{M}ulti \textbf{G}enerated \textbf{T}opic VAST (MGT-VAST) dataset, which extends VAST using GPT-3. In this dataset, each context is associated with multiple topics, allowing the model to understand the relationship between contexts and various potential topics
\footnote{The source code is available at \href{https://github.com/Babakbehkamkia/GPT-Stance-Detection}{https://github.com/Babakbehkamkia/GPT-Stance-Detection}}.

\end{abstract}

\section{Introduction}

With the growth of social media platforms, an increasing number of individuals turn to platforms such as Twitter for news consumption. Consequently, automatically identifying the opinions expressed in news articles and by people regarding specific topics (Stance Detection) has become a pressing issue in the field of Natural Language Processing (NLP) \cite{kaushal2021twt}. 
Initially, some researchers assumed that the topics encountered during the testing phase would align with those seen during training  \cite{mohammad2016semeval}. 

However, in reality, the number of topics is virtually limitless.
Unfortunately, collecting a comprehensive dataset that encompasses all possible topics is impractical. As a result, researchers have explored alternative approaches such as zero-shot and few-shot learning, as well as the extraction of topic-invariant features \cite{allaway-etal-2021-adversarial}. Nonetheless, none of these methods can outperform real data examples that explicitly convey the stance of a given post regarding a specific topic. 

In recent years, large language models (LLMs), like GPT-3 \cite{10.5555/3495724.3495883} and LLaMA \cite{touvron2023llama}, have brought a revolution to the NLP field. These models have been trained on large amounts of data and can understand and generate human-like text. Hence, LLMs hold significant potential for generating synthetic data that closely resembles real data.

By combining the concept of using LLMs with few-shot learning, we introduce a novel method for dataset generation. Additionally, to overcome the challenge of unseen topics in the testing phase, we suggest a novel approach. Our main contributions can be divided into two parts. 
 \begin{itemize}
     \item[--] We introduced MGT-VAST, a new dataset generated by GPT-3 from VAST \cite{Allaway2020Zero}, where each context is paired with multiple topics. The underlying idea is to assist the model in comprehending the relationship between a context and various possible topics. In comparison to the VAST Dataset, MGT-VAST contains a greater number of unique topics.
   
     \item [--] We propose the DyMoAdapt approach using LLMs in the test phase to enhance the performance of existing models. In this approach, we additionally fine-tune the model in the test phase with data  generated by GPT-3 according to the given topic. After fine-tuning, the model would be ready for the actual test data and perform better because it has seen similar examples of a particular topic.     
 \end{itemize}

\section{Related work}
Deep learning methods have proven to be effective in solving NLP problems, especially Stance Detection; however, they perform poorly when the training dataset is limited. Recently, few-shot learning has been proposed as a solution to this problem. This learning paradigm aims to generalize to new tasks with limited training data (zero or few labeled examples) using prior knowledge. The lack of labeled samples makes the estimation of the loss value during model training more challenging, which is the key issue of few-shot learning \cite{Hossain_2022}.
\newline
\citet{allaway-mckeown-2020-zero} is the first publication in the few-shot stance detection field. This research presents a new dataset called VAST. They collected this data according to the gap with existing datasets that contain a limited number of topics. They also introduced a novel deep learning approach that focuses on generalization when only a limited amount of data is available for each topic.
\newline
\citet{liu-etal-2021-enhancing} uses the VAST dataset and introduces a new model to improve its generalization capabilities. An enhanced general knowledge module was introduced to exploit semantic and structural level information. In this model, knowledge is limited to the relationships between documents and topics.
\newline
Another study introduced a zero-shot model called \textbf{TO}pic-\textbf{AD}versarial Network (TOAD), which employs adversarial learning \cite{allaway-etal-2021-adversarial}. They used domain-transfer ideas \cite{10.5555/3045118.3045244} to produce topic-invariant representations, allowing the model to generalize to unseen topics.
\newline
\citet{vamvas2020xstance} proposed a zero-shot model for generalizing across languages, in contrast to the previous work that focused on generalization across topics. They collected a dataset containing more than 150 political questions and 67k comments written by candidates. The comments comprise a mixture of German, French, and Italian. They fine-tuned a multilingual BERT model for stance detection.

\section{Methodology}
In this section, we first explain the concept of MGT-Vast data generation. Following that, we provide a detailed description of DyMoAdapt, the novel approach used during the test phase to address unseen topics.

\subsection{MGT-VAST Dataset}
In order to create the MGT-VAST dataset, we used GPT-3 to generate topics that are either in favor or opposing the existing posts in the VAST dataset. By using the prompt shown in Figure~\ref{topic-generation-propmt} The generated topics have lengths ranging from 2 to 4 words. Through this way, for each post, we have multiple topics, allowing the model to learn the relationship between each topic and the corresponding post. We have provided statistics for the MGT-VAST dataset, as shown in Table \ref{MGT-Vast-dataset-statistics} and Table \ref{MGT-Vast-dataset-mostfreq-topic}.

\begin{table}[ht!]
    \centering
    \begin{tabular}{c  |c c}
        \hline
         & Train& Test\\
        \hline
         \# Examples& 4986& 2305\\
         \# Examples with Agree label & 2516& 1204\\
         \# Examples with Disagree label & 2470& 1101\\
         \# Unique Post& 1233& 563\\
         \# Unique Topics& 4877& 2293\\  
         \# of Words in Topic & 17655 &8141\\
         \# of Unique Words in Topic & 5890 &3501\\
         Average \# of words per Topic & 3.51& 3.53\\
         \hline
        \hline
    \end{tabular}

    \centering
\caption{\label{MGT-Vast-dataset-statistics}MGT-VAST dataset Statistic}

\end{table}

\begin{table}[ht!]
    \centering
    \begin{tabular}{c  |c c c}
        \hline
         & Train& Dev& Test\\
        \hline
         SemEval-T6& 4870& - & 1956\\
         VAST& 13477& 2062& 3006\\  
         MGT-VAST& 4986& -& 2305\\
         \hline
        \hline
    \end{tabular}

    \centering
\caption{\label{dataset-statistics}This table illustrates the number of instances in each proportion of SemEval2016-T6, VAST, and MGT-VAST (our generated dataset).}

\end{table}
\begin{figure}[ht!]
  \centering
  \setlength{\unitlength}{1cm}
  \begin{picture}(10, 5)
    \put(0,0){
      \begin{tcolorbox}[
        arc=0mm,       
        width=7.5cm,     
        ]
        \raggedright 
        List the most potential topics of the given post with their labels. \\
        a label can be "agree" or "disagree" only. try to find both labels. \\
        summarize each topic in 2 to 4 words. \\
        return a JSON in which topics are keys and labels are values: \\
        example: \{\{\texttt{topic here}: \texttt{label here}\}\} \\
        post: ```\{\texttt{post}\}'''
      \end{tcolorbox}
    }
  \end{picture}
  \caption{\label{topic-generation-propmt}
  The prompt for dataset generation via GPT-3, which gets a post as input and gives all the possible topics for each post. }
\end{figure}
\subsection{DyMoAdapt Approach}
The intuition behind our proposed approach is to enhance the performance of existing models. This method operates as follows: For each new topic encountered during the test phase, we generate 2k new data points using GPT-3 (k samples in favor and k with opposing labels). We set k to 3 because of the limitations of the GPT\-3 API. Figur~\ref{generate-data-in-pipline-propmt} shows the prompt used to generate a synthetic post for the given topic. Afterward, the model is fine-tuned on the generated data. Then, with the fine-tuned model, predictions are made for the original test data. 
\begin{figure}[ht!]
  \centering
  \setlength{\unitlength}{1cm}
  \begin{picture}(10, 8)
    \put(0,0){
      \begin{tcolorbox}[
        arc=0mm,       
        width=7.5cm,     
        ]
        \raggedright 
        Your task is to generate a human written post. Do not 
        mention that you are an intelligent assistant.\\
        The generated post must discuss the given topic at some point in itself. \\
        The generated post must have a stance toward this topic. the stance could be "agree" or "disagree".
        the post should be at most 2 paragraphs.
        just return the post.\\
       topic: \{\texttt{topic}\}\\
        stance: \{\texttt{label}\}\\
        Here is an example post, but we do not know the stance of it toward the topic. you can learn from its structure.\\
       post: ```\{\texttt{post}\}```
      \end{tcolorbox}}
  \end{picture}
  \caption{\label{generate-data-in-pipline-propmt}
  The prompt for synthetic post generation via GPT\-3 in DyMoAdapt approach.}
\end{figure}
\begin{table*}[ht!]
\centering
\small
\begin{tabular}{p{4cm}|p{0.6cm}p{0.6cm}p{0.6cm}p{0.6cm}p{0.6cm}p{0.6cm}p{0.6cm}p{0.6cm}p{0.6cm}p{0.6cm}}
 \hline
\multirow{2}{=}{Model} & \multicolumn{4}{c}{VAST (\%)} & \multicolumn{6}{c}{SEM2016-T6 (\%)} \\
\cmidrule(lr){2-5}\cmidrule(lr){6-11}& Pro & Con & Neut & All & DT & HC & FM & LA & A & CC \\
 \hline
\hspace{2 mm}
\\
BERT & \textbf{82.3} & 65.3  & 16.68 & 37.2 & 33.5 & \textbf{57.6}  & 62.1 & \textbf{58.2} & 55.0 & 58.6 \\
\hspace{2 mm}
\\
TOAD & \centering{42.6} & \centering{36.7}& \centering{\textbf{43.8}}& \centering{\textbf{41.0}}&\centering{49.5} &\centering{51.2}  & \centering{54.1} & \centering{46.2}  & \centering{46.1} & 30.9 \\
\hspace{2 mm}
\\
GPT-3 & \centering{63.5} & \centering{69.1}& \centering{38.9}&\centering{40.4} &\centering{\textbf{53.7}}  & \centering{57.5} &\textbf{\centering{\textbf{62.3}}}  & \centering{47.9} & \centering{24.6} & 59.3 \\
\hline
\hspace{1 mm}
\\

DyMoAdapt (3 labels) & \centering{58.8} & \textbf{\centering{88.8}}& \centering{6.4}&\centering{35.8} &\centering{27.0}  & \centering{38.2} &\centering{21.2}  & \centering{29.3} & \centering{\textbf{58.9}} & \textbf{74.6} \\
\\
\\
 \hline
 \hline
\end{tabular}
\caption{\label{DyMoAdapt-approach-results}This table compares several baseline models with our pipeline.  We report $F1_{macro}$ for all of our experiments (the average of F1 on pro and con). It is important to note for BERT's experiments we only used a simple approach by just using the main text as input for this classification task. }

\end{table*}

\section{Experiments and Results}
Our experiments consist of two main parts: first, the evaluation of the MGT-VAST dataset, and second, the investigation of the DyMoAdapt approach. 

\subsection{MGT-VAST Dataset Evaluation}
For evaluating the MGT-VAST dataset, we chose three different models. Below, we describe the selected models and the experimental procedure:
\begin{enumerate}[label=\roman*]
    \item \textsc{BERT}: We selected BERT  \cite{devlin-etal-2019-bert}, which is a transformer-based model. The input is provided in the format "post [SEP] topic".
        \item \textsc{GPT-3}: LLMs that have recently been introduced and have demonstrated high performance in various tasks. In this experiment, we used the prompt shown in Figure~\ref{prompt-stance-detection} to instruct GPT-3 \cite{brown2020language} to determine the stance for the input post along with its topic. We utilized GPT3.5.Turbo and the OpenAI API.
    \item \textsc{TOAD}: another selected model is TOAD \cite{allaway-etal-2021-adversarial}, which utilizes Bicond LSTM\cite{augenstein-etal-2016-stance} and adversarial learning.
\end{enumerate}

\begin{table}[ht!]
    \centering
    \begin{tabular}{c  |c }
        \hline
         Topic& Frequency\\
        \hline
         Charter schools	& 6\\
         dual citizenship& 5\\
        Illegal Immigration & 4\\  
        Declawing cats & 4\\
        Immigration & 4\\
         \hline
        \hline
    \end{tabular}

    \centering
\caption{\label{MGT-Vast-dataset-mostfreq-topic}Most Frequent Topic in MGT-VAST (Train).}

\end{table}

The results obtained are displayed in Table~\ref{MGT-vast-result}. The analysis of the results on the MGT-Vast dataset demonstrates that the models have achieved promising outcomes. In particular, the BERT and GPT-3 models have outperformed TOAD in terms of  $F1_{macro}$. It is worth noting that the results of BERT and GPT-3 are quite comparable, and there is no significant superiority of one over the other in a meaningful sense. As expected, considering that a part of the MGT-VAST dataset has been generated using LLMs, transformer-based models have better performance.

\newcommand\tablespace{\rule{0pt}{2ex}}
\begin{table}[ht!]
\centering
\begin{tabular}{ p{1.6cm}|p{0.75cm}p{1cm}p{0.5cm}  }
\hline
\multirow{2}{=}{Model} & \multicolumn{3}{c}{Stance Label} \\
\cmidrule(lr){2-4} & Agree & Disagree & All \\
\hline
\tablespace
BERT & \centering{\textbf{68.5}} & \centering{\textbf{81.3}} & \textbf{60.0} \\
\tablespace
GPT-3 & \centering{68.4} & \centering{70.6} &  59.9 \\
\tablespace
TOAD & \centering{56.8} & \centering{47.5} &  52.2  \\
\hline
\end{tabular}
\caption{\label{MGT-vast-result} $F1_{macro}$ score of models on MGT-VAST dataset.}
\end{table}

 \begin{table*}[ht!]
\centering
\begin{tabular}{ p{10cm}p{4cm}p{1cm} }
\hline
post & Topics & Label \\\hline\hspace{1 mm}\\
\centering{\multirow{7}{=}{Without government to ensure their behavior, companies will attempt to make a profit even to the DETRIMENT of the society that supports the business. We have seen this in the environment, in finances, in their treatment of workers and customers. Enough.}} & Role of government & Agree \\
& Corporate behaviour & Disagree\\ 
& Profit motive & Disagree	\\
& Environmental impact & Agree \\
& Worker treatment & Agree \\
& Customer treatment & Disagree \\
& Social responsibility	 & Agree \\\\\hdashline\hspace{2 mm}\\
\multirow{8}{=}{I have two serious issues with plug-in cars, local and national. Locally, NH has the highest electric rates in the country (thanks, Seabrook). Nationally, plugging in is a huge waste of energy. To get a gallon of gas worth of electricity out of a wall socket, we need to put at least three gallons into the generator. Every time energy changes state, there is a loss: fuel to heat, heat to steam, steam to mechanical energy, mechanical to electrical, and in the car, electrical to battery, battery to mechanical, mechanical to tires, tires to motion. Add that to transmission losses, and you end up with a lot of waste instead of savings. Plug-in cars will add to the problem, not solve it.} & &  \\
& plugging in is a waste of energy & Agree\\\tabcolsep=10pt& &	\\& & \\
& plug-in cars will add to the problem & Disagree \\& & \\& & \\
& high electric rates & Disagree \\& & \\
\hline
\end{tabular}
\caption{In the first two rows, we demonstrate the generated topics and their stances towards the given text, while the last two instances are examples of generated posts related to a given combination of topic and label. We utilized GPT-3 to generate these posts, topics, and labels.}
\label{example}
\end{table*}
\subsection{DyMoAdapt Approach Evaluation}
 We chose BERT, GPT-3, and TOAD models to evaluate DyMoAdapt using the VAST\cite{allaway-mckeown-2020-zero} and SemEval2016-T6\cite{mohammad2016semeval} datasets. The procedure for obtaining the results of each model is explained as follows: 

\begin{enumerate}[label=\roman*]
    \item  \textsc{BERT}: we fined-tune BERT with a linear classification at the last layer on the training set for 10 epochs with a learning rate of 1e-5. Then, we evaluated the model on the test set of each dataset. The results are split through each topic (e.g., DT, HC, FM, LA, A, CC) for SemEval2016-T6 and for each stance label (e.g., Pro, Con, Neut) for VAST.
    \item \textsc{GPT-3}: we aimed to perform the stance classification task using prompts (Figure\ref{prompt-stance-detection}) with GPT3.5.turbo without fine-tuning due to our limited access to the OpenAI API.

    \item \textsc{TOAD}: the results of the TOAD model are from \citet{allaway-etal-2021-adversarial} research.

    \item \textsc{DyMoAdapt (3 labels)}: this experiment is similar to the BERT section in terms of training, with the main difference being in the testing phase. In the test phase, for each input, We asked GPT-3 to generate more posts according to the given topic using a prompt (Figure~\ref{generate-data-in-pipline-propmt}). Then, we fine-tuned BERT on the data generated by GPT-3 before performing the final predictions. This process is repeated for all samples in the test set.


\end{enumerate}

In general, the DyMoAdapt method can be a suitable alternative for use in real-time applications and with real data compared to other methods, including GPT and TOAD. The results obtained are presented in Table~\ref{DyMoAdapt-approach-results}.

\begin{figure}[ht!]
  \centering
  \setlength{\unitlength}{1cm}
  \begin{picture}(10, 5)
    \put(0,0){
      \begin{tcolorbox}[
        arc=0mm,       
        width=7.5cm,     
        ]
        \raggedright 
        What is the stance of the post which is delimited by 
       triple backticks toward the given topic? \\
      your answer should be agree, disagree or neutral.   keep your answer 1 word long. 
       please double-check your answer before responding and be sure about it.\\
       \textbf{topic}: \{\texttt{topic}\}\\
       \textbf{post}: ```\{\texttt{post}\}```
    \end{tcolorbox}}
  \end{picture}
  \caption{\label{prompt-stance-detection}
  The prompt for stance detection using GPT-3}
  
\end{figure}

\section{Discussion}
The data generated by GPT-3, which receives the neutral label, usually doesn't meet acceptable quality standards. Therefore, the performance of DyMoAdapt with three labels is weaker than DyMoAdapt with two labels. While the detection of the neutral label has consistently posed challenges for models, it is imperative to explore alternative methods for generating neutral data.

\section{Conclusion}
In this work, we proposed a new idea for the stance detection pipeline during the test phase called DyMoAdaptt, which almost improves performance on unseen topics. Using BERT with DyMoAdapt, achieved an average improvement of 24\% in F1 score across DT, HC, A, and CC topics in SEMEval2016-T6. However, it is important to note a corresponding reduction in performance for other labels.
Furthermore, we introduced the MGT-VAST dataset, which contains more than one topic for each post sample, generated using LLMs. The most important advantage of this dataset is the possibility of generating more samples.

For future work, Transformer-based models with attention layers can be employed to gain a deeper understanding of the relationship between topics and posts in the NLP domain. Moreover, other data augmentation methods, such as EDA can be used in DyMoAdapt to generate additional data and compare their results with the current approach.

\section{Limitations}
We generated topics with stances ("agree" or "disagree") towards a given text because we used GPT-3 to generate data, and GPT-3 couldn't generate an acceptable quality text with a neutral stance toward a topic. However, the stance detection task has three labels ("pro", "con", and "neutral"). Thus, our proposed dataset and pipeline don't perform well on instances with the "Neutral" label. Moreover, our experiments include a small portion of the datasets because we had very limited access to GPT-3.



\bigskip

\bibliography{aaai24}

\end{document}